# Designing a Virtual Manikin Animation Framework Aimed at Virtual Prototyping.

Antoine Rennuit[1, 2, 3], Alain Micaelli[1], Claude Andriot[1], François Guillaume[2], Nicolas Chevassus[2], Damien Chablat[3], Patrick Chedmail[3]

| 1: CEA\FAR – SRSI – LCI<br>18, route du panorama, BP 6<br>92265 Fontenay-aux-Roses<br>{micaellia;andriotc;rennuita}@<br>zoe.cea.fr | 2: EADS – CCR<br>12 rue Pasteur, BP 76<br>92152 Suresnes Cedex<br>{francois.guillaume;<br>nicolas.chevassus;<br>antoine.rennuit}<br>@eads.net | 3: IRCCyN<br>1, rue de la Noë, BP 92101<br>F-44321 Nantes Cedex 03<br>{Damien.Chablat;Patrick.Chedmail;<br>Antoine.Rennuit}@irccyn.ec-<br>nantes.fr |
|---|---|---|

**Abstract:** *In the industry, numerous commercial packages provide tools to introduce, and analyse human behaviour in the product's environment (for maintenance, ergonomics...), thanks to Virtual Humans. We will focus on control. Thanks to algorithms newly introduced in recent research papers, we think we can provide an implementation, which even widens, and simplifies the animation capacities of virtual manikins.*

*In order to do so, we are going to express the industrial expectations as for Virtual Humans, without considering feasibility (not to bias the issue). The second part will show that no commercial application provides the tools that perfectly meet the needs. Thus we propose a new animation framework that better answers the problem. Our contribution is the integration – driven by need - of available new scientific techniques to animate Virtual Humans, in a new control scheme that better answers industrial expectations.*

## Abbreviations

| | | |
|---|---|---|
| VH | : | Virtual Human |
| VP | : | Virtual Prototyping |
| DMU | : | Digital Mock-Up |
| CAD | : | Computer Aided Design |
| VR | : | Virtual Reality |
| PDM | : | Product Data Management |
| DOF | : | Degree Of Freedom |
| UC | : | Use Case |
| SDK | : | Software Development Kit |
| FK | : | Forward Kinematics |
| IK | : | Inverse Kinematics |
| FSM | : | Finite State Model |

## Introduction

The work we are introducing here takes place in a far larger context: the industrial design process. The Virtual Human (VH) is to fit the *concurrent engineering* design approach.

Large-scale concurrent engineering is now regarded as very attractive (one talk about "engineers' dream") [Dur03]. It is held by IT innovation as well as by new organisation and management methods. That is rational methods have to be developed to take advantage of industrial IT objects or machines, such as Digital Mock-Up (DMU) reviews to support large collaborative teams…

In such a framework, DMU is no longer an assembly model in a CAD tool, but an object managed by a Product Data Management (PDM) tool, which supports the product's integrity through collaborative work.

This collaborative aspect implies to control data flows: data exchange between people and IT machines, and data produced from other data thanks to people knowledge and software tools (knowledge management).

IT innovation for collective know how support (see figure below) is related to:
- *seamless virtual product* simulation and analysis, from early to in service models (Virtual Prototyping (VP))
- *technical IT data flow* from early investigation to downstream end users (data exchange)
- *knowledge cycles* from early concepts to knowledge support of end user (capitalize, and restore knowledge)

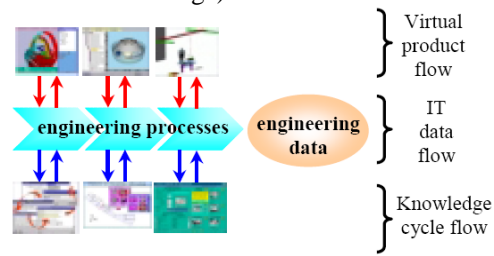

**IT support of collective know how in engineering process**

As stated above, this remains an "engineer's dream", several great problems still remain. The one we will focus on comes from obtaining a *seamless virtual product (red arrows on the above figure)*, and looks for *human centred design*.

A product is being designed to satisfy human's needs (Human Centric Design), thus human should naturally be incorporated within the design process. Actually, this is such a complex system, that it is very hard to take it into consideration as a constraint all along the design phase.

One should simulate, and introduce it into the VP virtual environment, but due to its high number of Degrees Of Freedom (DOF), controlling it is not so well handled: there are no thoroughly satisfying commercial solutions, and above all, they do not perfectly fit the VP framework. Need as for Virtual Humans is identified but the answers could be improved thanks to new control algorithms, and increasing computer capacities.

In the remaining sections of the paper, we will make a complete review of the industrial need, to understand better the industrial expectations as for Virtual Humans control. We will then establish a survey of the commercial packages available aimed at controlling a Virtual Human, show their deficiencies, and conclude to the inadequacy of existing solutions to the problem. This will allow us to introduce the new notion of *motion scale*, and thanks to it describe the design of a package better answering needs, thanks to newly introduced control algorithm.

## 1. Expressing need

As explained above, the virtual human, is a fantastic way to carry constraints linked to humans' interactions with the product (better known as ergonomics, accessibility checking...) in the designing process. A company using VH can expect benefits such as shorter design time, lower development costs, improved quality, increased productivity, enhanced safety, heightened moral... We can see VH fits perfectly to VP philosophy.

To fulfil all expectations of industrialists, we will adopt a systematic approach while expressing the need for a Virtual Human software package, in order to pick out the whole expression of need.

Ideally, the product will be tested in all situations. All these tests should be available to a VH animation package. To pick out them all, we analyse the product lifecycle.

In EADS, Product Lifecycle Management (PLM) analysis is seen as follows:

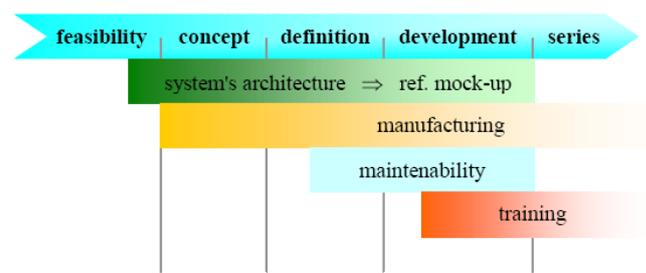

**PLM steps organisation in the concurrent engineering process, as seen by EADS. (simplified scheme highlighting the steps that will benefit most of VH)**

Improving the PLM steps' quality can directly be perceived as beneficial, and profitable by the client. It generates competitive advantages in a straightforward way!

In the following paragraphs, we will enumerate the stakes and challenges of each use case (which existed prior to the introduction of Virtual Prototyping), the contribution points Virtual Humans can bring to them, and then the competitive advantages resulting from using VH.

This will allow us to list all the needed features of the animation package to use, thus expressing our need.

### 1.1. Use cases

| VH in Operating |
|---|
| **Stacks & challenges:** |
| • develop human centred products, evaluate designs based on ergonomic factors |
| • account for diversity of people (sizes, shapes...) |
| • consider human factors before building physical prototypes |
| **VH contributions:** |
| • **positioning and comfort:** optimise user comfort, visibility, access to controls |
| • **visibility:** ensure differently sized people see what is important when manipulating the product |
| • **accessibility:** verify if the target population can easily climb in and out of the vehicle or equipment |
| • **reaching and grasping:** test if controls are placed in such a way that everybody can operate them, also consider foot pedal operations |
| • **multi-person interaction:** does the product fit collaborative work constraints? |
| • **emergency situations:** check evacuation, and crowd movements in case of emergency |
| • **strength assessment:** check if operating the product does not need extraordinary force, or create the potential for injury |
| **Competitive advantages:** |
| • *faster time to market:* VH allow more design |

| | |
|---|---|
| iterations in less time<br>• *higher product quality:* improved human centred design<br>• *reduced development costs:* digital mock-ups cost less than their physical counterparts (also takes less time to generate)<br>• *safer products:* more thorough analysis of user safety<br>• *improved productivity:* better automation of development process | • *mai*ntenance training: leverage computer technology to train maintenance personnel from multiple locations |
| | **Competitive advantages:**<br>• lower training costs<br>• more cost-effective training of geographically dispersed personnel<br>• heighten student retention of information<br>• improved workplace safety<br>• improved worker moral |

| **VH in Maintaining** |
|---|
| **Stacks & challenges:**<br>• reduce lifecycle costs by lowering maintenance requirements<br>• optimise maintainability easiness, and reduce downtimes<br>• ensure technicians can efficiently access parts, and manipulate tools needed for the task<br>• anticipate strength and time requirements for maintenance tasks |
| **VH contributions:**<br>• *reaching and grasping:* check if there is enough room for technicians to perform maintenance tasks, including space for tools<br>• *part removal and replacement:* ensure that all technicians can efficiently install and remove parts<br>• *visibility:* foresee what technicians can see when they perform a task<br>• *strength capability:* ensure it is possible, and not too difficult for a technician to perform its task. Reveal the need for collaborative work when needed<br>• *safety analysis:* be sure the technicians work in a safe environment |
| **Competitive advantages:**<br>• lower lifecycle costs<br>• less design rework and retrofit<br>• faster turnaround on maintenance jobs<br>• lower maintenance training costs |

| **VH in Manufacturing** |
|---|
| **Stacks & challenges:**<br>• improve user's safety<br>• reduce downtime and retraining costs<br>• improve product's efficiency |
| **VH contributions:**<br>• *workcell layout:* machines and equipments positions to optimise cycle times and avoid hazards<br>• *workflow simulation:* design manufacturing processes to eliminate inefficiencies and ensure optimal productivity. Simulate capabilities and limitations of humans to optimize the process<br>• *reaching and grasping:* check if workers can access parts, equipments, and manipulate the tools needed to perform the task<br>• *safety analysis:* ensure tasks are performed in a safe way<br>• *strength analysis:* check if manipulating the product does not need extraordinary efforts, or create the potential for injuries<br>• *energy expenditure:* calculate energy expended over time as workers perform a repetitive task, and optimise movement |
| **Competitive advantages:**<br>• fewer work related injuries, reduce workers' compensation costs<br>• more productive work environments<br>• improve employees' mood |

| **VH in Training** |
|---|
| **Stacks & challenges:**<br>• lower costs of training manufacturing and maintenance personnel<br>• train people without the need for physical prototypes or actual equipment: people could be trained before product availability, without monopolizing expensive equipments<br>• train people in multiple locations simultaneously |
| **VH contributions:**<br>• *manufacturing training:* use VR to train assembly workers on the virtual shop. Ability to modify reality to strengthen learning (ex: hide the blinding flash of lighting, when training welding, in order to see what we are doing)**.** |

Now that we showed the need for Virtual Humans, and revealed the benefits they could bring to industrialists, we will list all the features a VH animation package should incorporate to answer the whole need.

**1.2. Needed features**

The previous section allowed us to highlight all the contributions introducing VH into VP could bring.

We are now about to express the features needed by an animation package which aim is to support tools to bring all the promised contributions listed above (they were marked so as to see the direct relation between contributions and the features needed, ex: Ⓐ...). These previous contributions, now become requirements of the system to be built.

First we would like to stress on the nature of the current paper: it is "control" directed, and as such do not tackle the modelling of human bodies. This is another problem which is beginning to have smart, automatic solution, in particular in H. Seo, and N. Magnenat-Thalmann's papers [Seo03a], [Seo03b].

With hindsight from the contributions to be brought, we notice that the implications as for the animation system can be grouped into three main categories, which are the way we will specify the movement to be done, the way the system will be integrated within the existing VP scheme, and the tools the system has to implement. Thus we are going to follow this classification to express the requirements.

Note that, at the moment, we won't regard feasibility, but will just express the need as for the ideal animation system.

**Gesture/motion specification:**

What first appears when thinking about how to specify the movement, is the need for a completely autonomous system (see contributions A, B, C, D, E, G, H, I, J, M, N; P, Q, S, T), able to drive a Virtual Human through its environment in whatever situation, and thanks to *high level goals*. Simply think of a virtual worker, to whom you tell to screw one part of the product on another. This is a *high level goal*, because, it can be decomposed into smaller ones: reach his work station, grasp the screwdriver, the screw, put the screw into the hole, and finally screw. The decomposition depends on the situation, the environment. Moreover the goals are extremely numerous, and diverse. These points will be important when designing the new animation package.

Thanks to its adaptive[1] nature to environment, the autonomous Virtual Human perfectly fits the flexible concurrent engineering scheme. In concurrent engineering, process, maintenance, ergonomics, and other departments lead analysis at the same time. If a modification is done by a given department on the mock-up, it must not interact with the work done by other departments, or at least the other departments must be aware of the occurrence of the conflicting modification. Noticing that a modification can be conflicting, is often not so obvious, or does not come to mind quickly, above all when complex systems such as the human being are being considered. That is these conflicts are only revealed when tests are conducted... Thanks to autonomous VH, these tests can be achieved in an automatic way. Lets show an example: imagine a department wants to test the feasibility of maintenance route-sheets. Instead of simply conducting the tests, and analyse the results, the maintenance route-sheets will be saved, and the PDM automatically checks their integrity whenever a modification occurs.

This possibility of autonomous VH makes their true power: the VH's movement has to be reshaped for each modification, but the reshaping is automatic. It brings flexibility, and peace of mind to designers.

We have seen the power of autonomous controllers. But as for us, and without any technical consideration, less automated methods also present great interests. Actually, when conducting a design project, one often have to present the advances, show the remaining problems, discuss the different solutions in project reviews. To many people, automatically found solutions remain virtual until they have been tested.

One can go one step further in convincing decision makers, by making them try the product by themselves, immersing them in a virtual environment, and having them manipulate the product through their avatar driven by motion capture.

To us this solution must not be neglected, because it is much more persuasive than any other (technique useful in contributions C, D, G, ,H, I, M, Q, S).

Also note that we only talked about systems able to drive one VH. Certain tasks can not be achieved only by one man, and thus the system must be able to drive multiple VH. This plurality can be seen in two different ways.

On the one hand, there are cases where one has to drive simulations aimed at testing situations in which a task has to be achieved. Simply think of lifting an heavy load, this can require 2 or more men. In this case, work is well organised, we talk about *collaborative work*: that is several persons have a common goal they have to reach thanks to their respective skills, and capacities (L, R, T). This work group is looking for the synergy of the efforts of each.

On the other hand, there exists completely chaotic situations, where people only act on their own leading to muddled, disorganised, and inefficient movements (S). This is typically the case when crowds of people have to go through narrow passages, and its much more obvious in case of emergency. Think of a group of passengers rushing out of the plane at the same time, in case of incident.

---

[1] **Adaptive:** actually the VH is not strictly speaking adaptive, it does not adapt a movement to a situation, because the movement is not known a-priori, it is planed thanks to high level goals. If the environment changes, the planifier simply calculates another movement, without regarding previous movements.

Both kinds of movements are interesting to study when designing a product.

**Integration to VP scheme:**

Many of the contributions the VH promises to bring require analysing tools (A, B, D, E, F). Think of workflow simulation, or ergonomic analysis: these are domains which require measures to leaded and analysed. Of course these measures can be made on the Virtual Human, but, specialized packages are needed to help engineers analysis the results.

Actually animating Virtual Humans is the early stage of many different departments working on the product designed, and of many analysis: process, ergonomic, maintenance, tooling... Control is far from being an end in itself.

Most of these tools already exists, and it would be a nonsense to develop them again. That is we need to integrate the animation package in the existing VP scheme. To do so, we can not use a standalone software, which would prevent us from having other packages benefit by the animation package. We need to develop only control methods for manikins, that will allow us to drive VH in third party softwares.

The necessity for control methods instead of a standalone application could already be perceived through the text from the beginning, we simply express its need here.

**Related tools:**

In the contributions due to be brought by the Virtual Human, there are many tools related to VH that are quite obvious to obtain (C, L, P …).

As an example we could quote L, that can be checked through displaying vision cones, or viewing what the VH can see, or C which can be satisfied thanks to reaching envelopes, or measuring the distance between a hand and a point of the environment.

This kind of tools is very important when manipulating a VH, this is the kind of features which is handy, and brings a lot of information. However, they are very easy to obtain, and as such can be implemented in any animation package providing a Software Development Kit (SDK), if not already present in the package...

We will concentrate on the trickier aspects of the package, that is its animation capabilities.

Now we display a review of the commercial solutions available, and show their relative inadequacy to the problem.

## 2. Related work

In this section we will establish an overview of the solutions commercially available to animate a manikin in PLM's environment. The big issue at this point is to manage to obtain realistic movements (visually, physically, or better both of them), which satisfy the need we expressed earlier.

Surprisingly the technical solutions adopted in most packages are not very challenging. In most cases, softwares bring different solutions for the character to be animated, but the behaviour is left to the operator who completely drives the VH.

The simplest control scheme adopted is Forward Kinematics (FK)[2], thanks to this control the animator has full control of the movement, controlling each Degree Of Freedom, and movement is obtained thanks to *keyframing*. Unfortunately handling the manikin's posture using FK is not so easy, and is tremendously long.

So Inverse Kinematics (IK) is more often used, because it allows easier specification of postures: a reaching movement is directly specified in terms of IK target. This solution can be found in *eM-Human* from *Tecnomatix* [Tec], *Safework* [Saf], and *Man3D* [Man] (developed by *Renault*, a French car manufacturer, it is not commercially available)

Kinematics is quite easy to handle, unfortunately getting visually correct movements with kinematics is the domain of artists, and it is nearly impossible to achieve physically accurate ones, because it can not regard efforts. That is why another method is used: *dynamics*, which gives physical movements. It is nearly not currently used in commercial animation package, because of its high computational cost, except in Figure for Adams [Fig].

The technical solution that is mainly used for its simplicity, and its visual results, is to use a *library of motions* which can be retargeted, thanks to adequate algorithms, to the full size-range of VH. Blending the different movements, and making transitions between them can lead to very satisfying results. This technique is used in *DI-Guy* which better fits VH in a military context [Guy], *Ergo*, a module of *Deneb* [Erg], and *Jack* developed by University of Pennsylvania, and marketed by *EDS* [Jac].

The last method used in commercial packages is *motion capture*. It is used to widen the library of motions, or in a more powerful way to immerse directly the actor within the product's environment. This is a widely spread method: [Man], [Fig], [Erg]…

---

[2] **FK:** we assume the reader has a little knowledge about animation. Anyway, this light background can easily be found in any animation-related documentation.

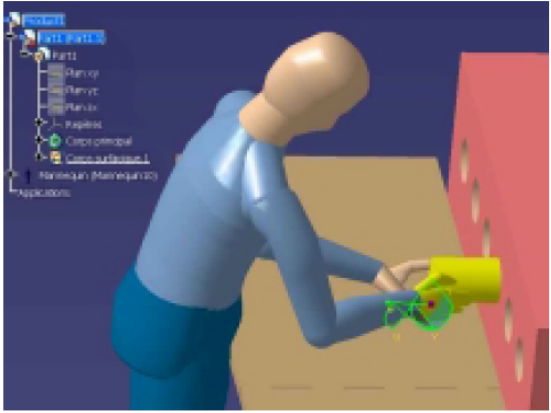

**Manikin under Delmia™ being controlled by IK.**

The packages available provide a given answer to the VH need. They all allow to test the product in its dedicated environment, they provide useful tools, and can answer many of the contributions listed above.

Unfortunately, as for us they have a restrictive vision of the possibilities, and completely miss important sides of VH.

When comparing the existing solutions to what we are looking for. We easily notice great divergences. Actually the only lack of autonomous control, for example, has great impacts. The most obvious one is the loss of time. Specifying a movement by hand, even when using keyframing with IK, or precomputed motion libraries, takes an enormous amount of time. Moreover getting a movement that does not look strange is a difficult task which constitutes a job itself, often found in entertainment animation companies. If there existed no other solution, we would be forced to deal with such techniques, but autonomous control for Virtual Humans is a reality.

One other aspect missed is it does not provide the DMU integrity checking as expressed above. The product cannot be tested as often as in the automatic case (because of the time it takes), therefore conflicting modifications can occur without being noticed early. We are moving away from the ideal concurrent engineering frame.

We shall also notice the near absence of certain pieces of the VH interest. Collaborative work, for example, is hardly considered, and the technique used to manage it, is the only juxtaposition of several simple controllers. Regarding the product to be handled with the workers that are to handle it, as a single system, brings a much richer set of solutions, than any other, since collaborative strategies are known to be much more effective. Furthermore, now, there exist techniques to answer such problems.

As for crowds, very little, if not nothing, is implemented in commercial packages. It seems this side of VH remains the domain for computer graphics for entertainment (mainly for movie), or it is confined to projects of very special interest like renowned buildings such as "Le Stade de France", a great stadium near Paris – although it is not exactly the field of application which interests us most.

Most of the techniques used in the existing softwares are very traditional. This is correlated to their age: when having a look at when they where carried out, we can see they are not so new, and thus do not benefit by newly introduced algorithms.

This diagnostic pushed us to regard new methods. The approach we will now introduce gain by them. We expressed what we expected so as to virtual manikins, and thanks to state of the art techniques, we will be able to design a complete alluring answer to need.

## 3. Designing the animation system

We are now about to describe a new animation strategy which will take us much further in animation systems.
Need is expressed, we are now to answer it.

### 3.1. Motion scales

Until now, we talked about the need for automation, high level goals, and other features, without regarding any technical constraint. They exist...

When looking at a worker during his work, we notice the great panel of movements he has to do: manipulating parts, assemblies, moving between the different manufacturing cells of his workshop, or larger movements following the flow line, coordinated movements for collaborative work…

Depending on the system to animate, and to what we want it to do, technical solutions will be different.

In order to formalize the problem, we introduce the notion of *motion scale* which roughly stands for the type of animation, and the complexity of the system we want. Analysing the motion scale's space will help us identify the technical solutions that will solve our problem.

As a rough introduction to motion scales we will describe examples:
- manipulating a part is done only by the hand, it is low motion scale
- manipulating assemblies is made thanks to upper extremities[3], it is an higher motion scale
- displacement from one location to another imply the animation of the whole body, it is an higher scale

---

[3] **Upper extremities:** both hands, arms, & trunk

- certain tasks are more complex, they are compositions of simpler movements, it is yet an higher scale
- the system is even more complex: moving an heavy load requires several human

The notion of motion scale still must appear a bit bewildering. Several aspects are mixed together: the complexity of the skeleton to animate, the length of displacements, the number of VH to animate… This is because motion scales live in a multidimensional space.

The first one is the complexity of the system. In fact, this dimension is dual. If we quantify the complexity of the system thanks to the number of DOF to animate, we obtain the following scheme:

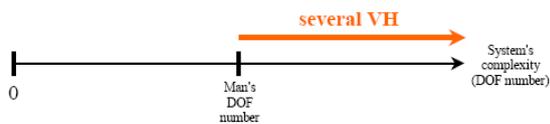

**Complexity dimension splits into two components.**

That is on the one hand we deal with the VH's *skeleton's complexity* (shall we consider hands, legs, simplifications with respect to the actual human skeleton…), on the other hand we regard if we introduce *several manikins*. The techniques to answer these two problems are completely different, furthermore, the matter of skeleton's complexity is independent from the one of the number of VH, therefore we differentiate the two dimensions.

The last component is the *spacio-temporal* dimension of the control. A simple local movement aimed at moving into position a part with the only arm does not have the same nature as a complex movement implying composition of large scale displacements (reaching a location can imply walking, and climbing a ladder for example). We have to account for the plurality of solutions this dimension implies, thus we integrate it to the motion scale' space.

Depending on the position of our movement in the motion scale's space, different technical solutions will be used to animate the VH.

### 3.2. Control framework

We know that depending on the purpose, we will be using different technologies. As explained above automation will be used for the integration into the concurrent engineering scheme, and motion capture is an option that must be available for reviews.

We should add that due to technical constraints movements obtained through motion capture are limited to small portions of space (observation surfaces are very limited, whatever the motion capture technique used), however, the movements obtained are more natural-looking than with automated methods (which can handle any motion).

The different algorithms available do not reach the same parts of the motion scale' space:

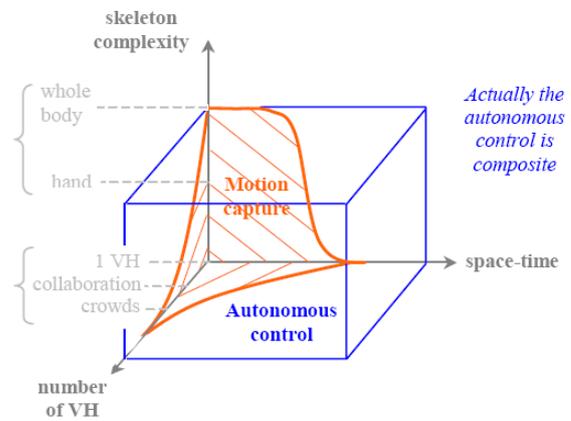

**Motion scale' space.**

We see the autonomous control can render any motion. Actually, mainly due to the high number of DOF of the human being, there do not exist any controller that can handle the whole motion scale' space. Moreover, we have the same problem with motion capture which – at least at the moment - can not be dealt with only one device. We will have to compose controllers.

#### 3.2.1. Solution to simplified problem

To make things easier, we will first project the problem on the plan spanned by *skeleton complexity*, and *space-time*, for only one Virtual Human, and then extend the solution to the whole motion scale' space.

**Space-time compositing:**

At this stage, we will compose low-level controllers (watching out simple movements[4]), thanks to a high-level controller which will manage the transition between low level ones.

Low-level controllers already exists [Kuf03], or can be created on purpose. On must find the high-level controller.

This high-level framework already exists, P. Faloutsos, M. van de Panne, and D. Terzopoulos created a system able to composite dynamic controllers, while taking care of balance in a very smart way [Fal01], & [Fal03].

Individual controllers are seen as black boxes, which means that providing they are dynamical,

---

[4] **Simple movements:** walking, climbing a ladder, crawling, arm reaching an object, grasping are said to be simple movements. Complex movements are compositions of these simple movements.

controllers can be whatever, the only constraint is one must be able to determine pre-conditions[5], post-conditions[6], and performances[7]. The complete sequence is as follows:
- default position
- user invokes a series of simple movement: the high-level controller chooses the best low level-controller able to accomplish the first movement asked
- when accomplished, the high-level controller chooses the best controller able to accomplish next movement, and so on…
- if balance is disturbed because of an unpredicted event (imagine the manikin is hitten by a falling object), then the high level controller chooses the best low-level controller to recover balance if possible, otherwise it chooses the best low-level controller to fall

A. Shapiro, and P. Faloutsos later introduced an extension to the framework to manage hybrid kinematic / dynamic low-level controllers [Sha03]: that is nearly all controllers can be combined in the same framework. Also notice the toggle between motion capture and autonomous control can be handled by this hybrid high-level controller.

At this stage of the paper, we are able to control simple motions, and to compose them into more complex movements. We now have to tackle the sequencing of controllers: choosing the right sequence of right controllers at the right time to obtain the desired high-level behaviour of our manikin. If we tell a VH to remove a part, it must be decomposed in reaching the location of the part, grasping the tool, and removing the part. Moreover, depending on the situation, the sequencing can be different, sharper, and can evolve during the simulation itself according to the surrounding environment's evolution, that is the VH should be reactive[8]…

This is a very wide problem. The evolution of the VH can be manually driven by an operator, but it would prevent the appeal of automatic control.

D. Thalmann's VrLab developed a behaviour model based on combination of rules, and Finite State Machines (FSM) which can guide VH to high-level goals in an adaptive way [Tha02]. The FSM provides the behaviour for pre-established standard high-level goals, and the rules tell how to behave when given events occurs.

**Skeleton compositing:**

We saw how dubbing of controllers along time could help us achieve complex movements. We are now to tackle another type of complexity: the one of skeleton. This is mainly useful for motion capture, but can still be interesting for "exotic" automatic controllers.

When having a look at the commercial devices that allow us to capture motion, we see that none of them can cope with the whole body at once. We have devices for hands such as data gloves, and other devices that allow to capture the remaining parts of the body. If we want movement observation of the whole body including hands, we have to combine the different sources of information.

Kinematical, or dynamical mixing of movements do not raise particular problems. Troubles arise when regarding hybrid kinematical / dynamical combination of movements. That is we have a kinematical controller for the hands, and want a dynamic behaviour of the body (the inverse case does not show such difficulties). In this case, the kinematical movement of the hand has an influence on the dynamical motion of body, which has to be taken into account; it is not obvious to find because of the heterogeneous nature of controls.

Once again, there already exists an algorithm to deal with such problems. Isaacs, and Cohen introduced in [Isa87] a general scheme allowing to dynamically combine complementary motions from different sources. The kinematical joint motion of given parts of the skeleton is given as input of the system which is dynamically solved for the remaining joints.

Two dimensions of the motion scale' space are now covered, next section is about the last one: multiple VH.

### 3.2.2. Extension to several VH

The techniques used to animate several VH at the same time can rely on the ones seen to animate a unique VH. However, one have to manage the interactions between these humans. As explained in previous sections this interaction can be found under two shapes: organised interactions in collaborative tasks, disorganised interactions in the case of crowds.

This is somewhat a behaviour level, since we decide the movement the VH will achieve.

---

[5] **Pre-conditions:** initial set of the states of the manikin to be controlled in which the controller is able to cope with the movement.
[6] **Post-conditions:** set of final states of the manikin reachable by the controller
[7] **Performance:** when the system has to choose between different possible controllers, it chooses the best one, that is why we have to evaluate performance
[8] **Reactive VH:** this reactiveness is a very general concept, it also includes communication with other manikins [Tha02].

**Collaborative VH:**

Collaborative manipulation of an object is quite demanding, it requires a strategy to take advantage of the synergy of the means available, and then being able to control the new system which is much more complex than in previous case.

To control the system, we lean on O. Khatib's *augmented object model* [Kha95], which describes the dynamic evolution of an object manipulated by several contact points (several hands). Khatib shows that the augmented object as an equation of motion that has the same shape as the one of a single manipulator, thus the control techniques to be used can be the same as the well known ones of single manipulators [MLS95].

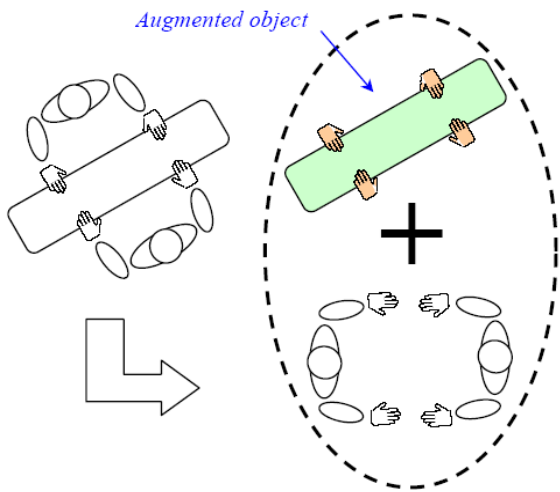

**Augmented object: the system being considered is the held object.
VH are controlled in a second step.**

Notice that augmented objects can be found in nearly all manipulation cases: an object manipulated with both arms of a single human, can be considered as an augmented object, since there are multiple contacts. An object manipulated with several fingers can also be seen as an augmented object[9]…

**Crowds:**

Simulating crowds is a completely different problem in essence, mainly because of the muddled nature of crowds.

D. Thalmann has an interesting approach to crowds control [Tha02]. He focuses on complex behaviours of many agents in dynamic environments

---

[9] **Augmented object:** regarding all DOF of the manikins would lead to huge augmented objects systems: most often we can not consider the dynamic of the whole augmented system. One strategy to deal with such cases is to study the hands separately from the rest of the manikins' bodies.

with possible user interaction. Unlike other approaches [Hel00], here crowds are dynamically assembled and disassembled, and over time they change their behaviour. Member of a gathering do not act as a whole, they operate in subsections which react in similar ways, this emerges in a group behaviour has sociologists teach it.

Other agents and real human participants (through motion capture) of the simulation, can their own behaviours, and interact with the crowd, which greatly widens the possibilities of the system.

The techniques described above allow us to span all the motion scale' space, next comes a recapitulative diagram of the whole system.

### 3.3. Virtual Humans' control scheme

When looking at all the functions needed to implement our VH animation approach, we differentiate two classes: the ones that plan the movement and the manikin's behaviour, and the ones that allow to carry out, and control it[10].

Here is the scheme of the whole system:

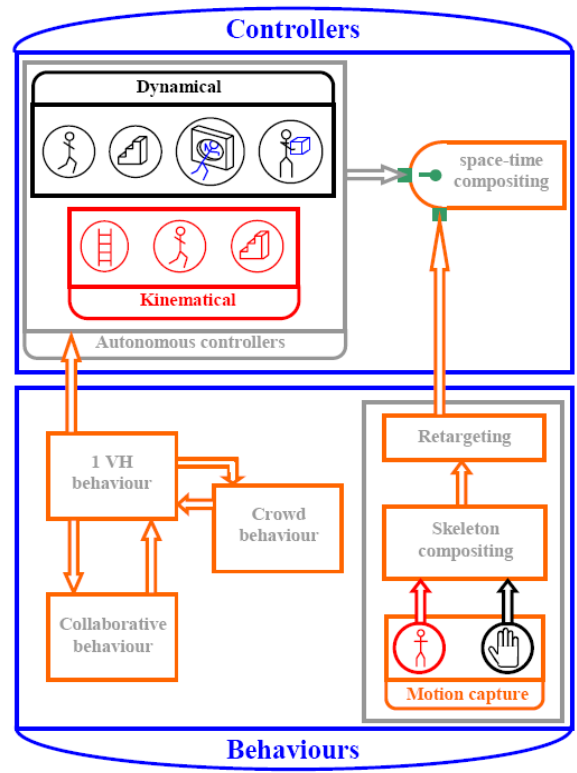

**Whole animation framework**

---

[10] **Behaviour / control:** There exist a direct identification with cervical areas of a real human: behaviour is mainly achieved by the *motor cortex*, and control is achieved by the *cerebellum*.

This approach is very flexible, because, it can follow the quick evolution of the scientific community: if a new controller is created, it can be easily implemented in our framework. And it is the same for any function in the system: this is the power of a modular approach.

## Conclusion

In this paper, we first reminded the context of our work, that is concurrent engineering, and VP, taking advantage of IT capacities. Then we focused on VH animation : first we analysed, and expressed needs as for VH within the VP context. Then we made a survey of the techniques used in commercial packages to animate VH, and showed their relative inadequacy to the problem. That led us to carry out the specification of a new control scheme based on the integration of algorithms recently introduced in the scientific community, and whose architecture was driven by need.
We think this new framework better meets the needs than previously existing ones, and is more flexible, and evolutive.

## Bibliography

**[Dur03]:** "Collaborative large engineering: from IT dream to reality", Michel Dureigne, .

**[Erg]:** www.deneb.com/products/ergo.html

**[Fal01]:** "Composable Controllers for Physics-based Character Animation", Petros Faloutsos, Michiel van de Panne and Demetri Terzopoulos, *Proceedings of ACM SIGGRAPH 2001*, Los Angeles, August 2001.

**[Fig]:** www.adams.com

**[Fal03]:** "Autonomous Reactive Control for Simulated Humanoids" , Petros Faloutsos, Michiel van de Panne and Demetri Terzopoulos, *IEEE International Conference on Robotics and Automation (ICRA) 2003*, Taiwan, pp. 917-924.

**[Guy]:** www.bdi.com

**[Hel00]**: "Simulating dynamical features of escape panic", D. Helbing, I. Farkas, T. Vicsek, Nature 407, pp. 487-490, 2000.

**[Isa87]:** "Controlling Dynamic Simulation with Kinematic Constraints, Behavior Functions and Inverse Dynamics", P. Isaacs, M. Cohen, *Proceedings of ACM Siggraph 1987*.

**[Jac]:** www.plmsolutions-eds.com

**[Kha95]:** "Inertial Properties in Robotic Manipulation: An Object-Level Framework", Oussama Khatib, *International Journal of Robotics Research*, Vol. 14, N° 1, February 1995, pp. 19-36.

**[Kuf03]:** "Motion Planning for Humanoid Robots", J. Kuffner, K; Nishiwaki, S. Kagami, M. Inaba, H. Inoue, *11 th International Symposium of Robotics Research* (ISSR 2003).

**[Man]:** "Man 3D: un mannequin numérique pour la simulation ergonomique", J.P. Verriest, *Humanoid Day,* Societe Francaise de Biomecanique, 15 mai 2003, Valencienne.

**[MLS95]:** "A Mathematical Introduction to Robotic Manipulation", R. Murray, Z. Li, S. Sastry, Published by CRC Press, 1995.

**[Saf]:** http://www.safework.com/

**[Seo03a]:** "An Automatic Modeling of Human Bodies from Sizing Parameters", H. Seo, N. Magnenat-Thalmann, *ACM SIGGRAPH 2003 Symposium on Interactive 3D Graphics*, pp. 19-26, pp234, 2003.

**[Seo03b]:** "Synthesizing Animatable Body Models with Parameterized Shape Modifications", H. Seo, F. Cordier, N. Magnenat-Thalmann, *ACM SIGGRAPH / Eurographics Symposium on Computer Animation*, pp. 120-125, July, 2003.

**[Sha03]:** "Hybrid Control For Interactive Character Animation", Ari Shapiro, Frederic Pighin, and Petros Faloutsos, *Pacific Graphics, 2003*, short paper, pp. 455-461.

**[Tec]:** www.tecnomatix.fr/showpage.asp?page=405

**[Tha02]:** D. Thalmann, Simulating a Human Society: the Challenges, *Proc. CGI 2002*.